\documentclass{article}

\usepackage{graphicx}
\usepackage{afterpage}
\usepackage{booktabs} 

\usepackage{hyperref}

\usepackage{lipsum}
\usepackage{url}

\usepackage{breakurl}
\usepackage{caption}
\usepackage{subcaption}

\usepackage{amsmath,amsfonts, amssymb}
\usepackage{multicol}
\usepackage{multirow}
\usepackage{array}
\usepackage{mathtools}
\usepackage{epstopdf}
\usepackage{amsthm}
\usepackage{bbm}
\usepackage{dsfont}
\usepackage{microtype}

\newcolumntype{P}[1]{>{\centering\arraybackslash}p{#1}}
\newcolumntype{M}[1]{>{\centering\arraybackslash}m{#1}}

\usepackage[accepted]{mlsys2020}
\begin{document}

\twocolumn[
\mlsystitle{Distributed Sparse SGD with Majority Voting}




\begin{mlsysauthorlist}
\mlsysauthor{Kerem Ozfatura}{ozu}
\mlsysauthor{Emre Ozfatura}{imp}
\mlsysauthor{Deniz G\"und\"uz}{imp}
\end{mlsysauthorlist}

\mlsysaffiliation{imp}{Information Processing and Communications Lab, Dept. of Electrical and Electronic Engineering,Imperial College London, London, UK}
\mlsysaffiliation{ozu}{Department of Computer Science, Ozyegin University, Istanbul, Turkey}

\mlsyscorrespondingauthor{Kerem Ozfatura}{kerem.ozfatura@ozu.edu.tr}

\mlsyskeywords{Machine Learning, MLSys}

\vskip 0.3in

\begin{abstract}
Distributed learning, particularly variants of distributed stochastic gradient descent (DSGD), are widely employed to speed up training by leveraging computational resources of several workers. However, in practise, communication delay becomes a bottleneck due to the significant amount of information that needs to be exchanged between the workers and the parameter server. One of the most efficient strategies to mitigate the communication bottleneck is top-$K$ sparsification. However, top-$K$ sparsification requires additional communication load to represent the sparsity pattern, and the mismatch between the sparsity patterns of the workers prevents exploitation of efficient communication protocols. To address these issues, we introduce a novel majority voting based sparse communication strategy, in which the workers first seek a consensus on the structure of the sparse representation. This strategy provides a significant reduction in the communication load and allows using the same sparsity level in both communication directions. Through extensive simulations on the CIFAR-10 dataset, we show that it is possible to achieve up to $\times4000$ compression without any loss in the test accuracy.
\end{abstract}
]


\printAffiliationsAndNotice{}  

\section{Introduction}
Advances in the design of neural network (NN) architectures exhibit impressive results in many challenging classification problems \cite{NN.VGG, NN.DRL, NN.DRN, NN.DenseNet}. However, these designs mainly promote deeper NN structures, which come at the cost of significant training time, as they also require much larger datasets for training. As a result, training these deep NNs (DNNs) has become a formidable task that is increasingly infeasible to perform on a single machine within a reasonable time frame. The solution is to employ a distributed training framework, where multiple machines/workers are used in parallel to reduce the training time under the supervision of a parameter server \cite{PSGD1,PSGD2,PSGD3,PSGD4}.

\indent From a theoretical point of view parameter server type implementations, such as synchronous parallel stochastic gradient descent (SGD), offer a speed up of training that scales linearly with the number of workers. However, such a speed up is typically not possible in practise as distributed training requires exchange of large amounts of information between the parameter server and the workers, and the communication latency becomes the bottleneck. The communication bottleneck becomes particularly pertinent when the communication is performed over bandwidth limited channels, for example, when the workers are edge devices \cite{FL_1, FL_2}. Numerous techniques have been proposed in the recent years to reduce the communication load in distributed learning. Next, we briefly explain the general distributed SGD (DSGD) framework, and then briefly overview some of the popular strategies to reduce the communication load.

\subsection{Preliminaries}
Many parameterized machine learning problems can be modeled as a {\em stochastic optimization problem}
\begin{equation}\label{problem}
\min_{\boldsymbol{\theta}\in\mathbb{R}^{d}} \mathrel{\mathop:}= \mathds{E}_{\zeta \sim \mathcal{D}}F(\boldsymbol{\theta}, \zeta),
\end{equation}
where $\boldsymbol{\theta}\in\mathbb{R}^{d}$ denotes the model parameters,  $\zeta$ is the random data sample, $\mathcal{D}$ denotes the data distribution, and the $F$ is the problem specific empirical loss function. Stochastic optimization problem given above can be solved in a distributed manner over $N$ workers by rewriting the minimization problem in (\ref{problem}) as
\begin{equation}
\min_{\boldsymbol{\theta}\in\mathbb{R}^{d}} f(\boldsymbol{\theta})= \frac{1}{N}\sum^{N}_{n=1}\underbrace{\mathds{E}_{\zeta \sim \mathcal{D}_{n}}F_{n}(\boldsymbol{\theta},\zeta)}_{\mathrel{\mathop:}=f_{n}(\boldsymbol{\theta})},\label{DSO}
\end{equation}
where $\mathcal{D}_{n}$ denotes the portion of the dataset allocated to worker $n$. Most machine learning algorithms employ SGD for the solution of the above problem. In PS-type implementation of DSGD, at the beginning of 
iteration $t$, each worker pulls the current global parameter model $\boldsymbol{\theta}_t$ from the PS and computes the {\em local stochastic gradient}
\begin{equation}
\mathbf{g}_{n,t}=\nabla_{\boldsymbol{\theta}_{t}}F_{n}(\boldsymbol{\theta}_{t},\zeta_{n,t}),
\end{equation}
where $\zeta_{n,\tau}$ is the sampled training data at iteration $t$ by the $n$th worker. Then, each worker sends (pushes) the value of its local gradient estimate to the PS, where those values are aggregated to update the parameter model, i.e.,
\begin{equation}\label{model_update}
\boldsymbol{\theta}_{t+1}= \boldsymbol{\theta}_{t}-\eta_{t}\underbrace{\frac{1}{N}\sum^{N}_{n=1}\mathbf{g}_{n,t}}_{\mathbf{G}_{t}},
\end{equation}
where  $\eta_{t}$ is the learning rate.

\subsection{Communication efficient distributed learning}
\indent There is a plethora of works on communication efficient distributed learning, which can be classified into three main categories; namely, sparsification, quantization, and local SGD. Let $\mathbf{g}$ denote the generic gradient estimate to be transmitted from a worker to the parameter server at some iteration of the DGSD algorithm. 

\subsubsection{Quantization}
The objective of the quantization strategy is to represent $\mathbf{g}$ with as few bits as possible. Without quantization, each element is represented with 32 bits according to the floating point precision. The typical approach is to employ scalar quantization, where each value of $\mathbf{g}$ is quantized into fewer than 32 bits.  In the other extreme, each element of $\mathbf{g}$ can be represented by a single bit \cite{SGD.q4, SGD.q5}. Numerous quantization  strategies have been studied in the literature for the DSGD framework \cite{SGD.q0, SGD.q1, SGD.q2, SGD.q3, SGD.q6}.

\subsubsection{Sparsification}
The core idea behind sparse communication is to obtain a sparse version of the vector $\mathbf{g}$, denoted by $\tilde{\mathbf{g}}$, such that at most $\phi$ portion of the values in $\tilde{\mathbf{g}}$ are non zero, i.e.,
\begin{equation}
\vert\vert\tilde{\mathbf{g}}\vert\vert_{1} \leq \phi \cdot d,
\end{equation}
where $d$ is the dimension of $\mathbf{g}$. Then, only the non-zero values in $\tilde{\mathbf{g}}$ are transmitted to the parameter server. In general, $\tilde{\mathbf{g}}$ is represented using a sparsity mask $\mathbf{M}\in \left\{0,1\right\}^d$, such that $\tilde{\mathbf{g}} = \mathbf{M}\otimes \mathbf{g}$, where $\otimes$ denotes element-wise multiplication. The key challenges for sparse communication is the design of the sparsity mask $\mathbf{M}$, and the representation of the positions of its non-zero entries with minimum number of bits. Recent works have shown that employing sparse communication with $\phi\in[0.01,0.001]$ while training DNN architectures, such as ResNet or VGG, reduces the communication load significantly without a notable degradation in the accuracy performance \cite{SGD.sparse1, SGD.sparse2, SGD.sparse3, SGD.sparse4, SGD.sparse5, SGD.sparse6,SGD.sparse7, SGD.sparse.rtopk, SGD.sparse.FL}.\\

\subsubsection{Local SGD}
Another common strategy to reduce the communication load is local SGD. While the previous two strategies aim to reduce the number of bits sent at each communication round, local SGD aims to minimize the total number of communication rounds by simply allowing each worker to update its model locally for $H$ consecutive SGD steps \cite{SGD.local1, SGD.local2, SGD.local3, SGD.local4, SGD.local5, SGD.local6, SGD.local7,SGD.local8,SGD.local9}. This implies $\times H$ reduction in the number of communication rounds compared to standard DSGD. Local SGD 
mechanism is also one of the key features of the popular federated learning framework \cite{FL1}.

\subsection{Motivation and Contributions}
For communication efficient distributed learning, one of the most efficient and widely used strategies is the top-$K$ sparsification framework, especially when it is implemented together with the error feedback/accumulation mechanism. However, there are certain drawbacks of the top-$K$ sparsification framework. First, although sparsification reduces the number of parameter values to be transmitted at each communication round, it requires the transmission of additional bits to identify the positions of non-zero entries in sparse representations. Second, since the positions of the top-$K$ values do not necessarily match across the workers, top-$K$ sparsification is not a linear operation. However, since the ultimate aim at the parameter server is to compute the average of the gradient estimates over all the workers, instead of receiving each one separately, linearity property can help to employ more efficient communication strategies both in wired \cite{SGD.sparse0} and wireless network setups \cite{oa2,oa1,oa3}. Finally, another disadvantage of the mismatch between the top-$K$ positions among the workers is that the promised  sparsification level is only satisfied at the uplink direction.

\indent To overcome these limitations, we introduce a majority voting based sparse communication strategy building upon the top-$K$ sparsification framework with an additional consensus mechanism, where all the workers agree on the structure of the sparsity. Furthermore, we show that by establishing a certain correlation between the non-sparse positions of the gradient estimates observed in consecutive iterations, the proposed voting scheme can be modified to utilize this correlation and achieve a lower communication load, measured in terms of the total number of transmitted bits, which is critical when the quantization and sparsification strategies are employed together.

\indent Beyond communication efficiency, we also want to establish a relationship between the sparsity of the NN architecture obtained as a result of the distributed training process, and the sparsity used for communication. In other words, we explore if a tailored sparsification strategy for communication can exploit the sparse nature of the NN architecture, similarly to network pruning \cite{NN.prune1,NN.prune2,NN.prune_dynamic}. Indeed, as further highlighted in Section \ref{section:numerical}, we observe that sparse communication can also result in improved accuracy results; hence, we argue that a sparse communication strategy that is aligned with the sparse nature of the NN architecture may help to achieve better generalization.

\section{Sparse Communication with Majority Voting}
One of the most commonly used techniques for communication efficient DSGD is top-$K$ sparsification. Let $\mathbf{g}_{n,t}$ denote the local gradient estimate computed by the $n$th worker at iteration $t$. The worker sends a sparse version of its gradient estimate $\mathbf{g}_{n,t}$, denoted by $\widetilde{\mathbf{g}}_{n,t}$, by keeping only the $K$ largest absolute values. That is,
\begin{equation}
\widetilde{\mathbf{g}}_{n,t}=\mathbf{g}_{n,t}\otimes \mathbf{M}_{n,t},
\end{equation}
where $\mathbf{M}_{n,t}=S_{top}(\mathbf{g}_{n,t},K)$ is the sparsity mask used by the $n$th worker. Here, $S_{top}(\mathbf{g},K)$ returns a vector with $K$ ones corresponding to the $K$ elements of vector $\mathbf{g}$ with the highest absolute values. Accordingly, the PS aggregates the sparse gradients  $\widetilde{\mathbf{g}}_{1,t},\ldots\widetilde{\mathbf{g}}_{N,t}$ from the $N$ workers to obtain the global gradient estimate $\widetilde{\mathbf{G}}_{t} \triangleq \frac{1}{N} \sum_{n=1}^N \widetilde{\mathbf{g}}_{n,t}$, which is then used to update the global model, i.e.,
\begin{equation}
\boldsymbol{\theta}_{t+1}=\boldsymbol{\theta}_{t} + \eta_{t} \widetilde{\mathbf{G}}_{t}.
\end{equation}
The key design objective of top-$K$ sparsification is to reduce the communication load in the uplink direction, i.e., from the workers to the parameter server; thus the sparsification levels in the uplink and downlink directions do not necessarily match, and different sparsification masks can be employed. 

\indent  We remark that if all the workers use the same sparsity mask, i.e., $\mathbf{M}_{n,t}=\mathbf{M}_{t}$, $\forall n$, then the sparsification becomes a linear operation; that is,
\begin{equation}
\widetilde{\mathbf{G}}_{t}=\mathbf{G}_{t}\otimes\mathbf{M}_{t}= \frac1N \sum^{N}_{n=1}\mathbf{g}_{n,t}\otimes\mathbf{M}_{t}=
\frac1N \sum^{N}_{n=1}(\mathbf{g}_{n,t}\otimes\mathbf{M}_{t}).
\end{equation}
This linearity  property can be utilized to employ more efficient collective operations in the distributed setup, such as using {\em all-reduce} operation instead of {\em all-gather} \cite{SGD.sparse0}, or to take advantage of the superposition property of wireless channels \cite{oa1,oa2,oa3} when the workers are edge devices.

\indent In addition to aforementioned discussion on efficient  sparse communication, we also argue that using a common sparsity mask, besides reducing the communication load, can also allow exploiting the sparse nature of the deep NN architectures; and hence, achieve better generalization and higher test accuracy. Later, by conducting extensive simulations, we show that inducing sparsity with a certain structure for communication can indeed help to achieve better generalization performance.

\indent In this work, we want to design a computationally efficient  mechanism to construct a global sparsity mask $\mathbf{M}_{t}$ at each iteration $t$. Inspired by \cite{SGD.q5}, we introduce a {\em majority voting (MV)} based mechanism to construct the global sparsity matrix, which, in a broad sense, corresponds to jointly identifying the most important $K$ locations for the global model update across all the workers.

The voting mechanism is executed in the following way: at each iteration $t$, each worker computes the local gradient estimate $\mathbf{g}_{n,t}$, and identifies the largest $K$ absolute values $\mathbf{M}_{n,t}=S_{top}(\vert\mathbf{g}_{n,t}\vert,K)$, and sends the locations of the non-zero entries of $\mathbf{M}_{n,t}$, denoted by $\mathrm{supp}(\mathbf{M}_{n,t})$, to the PS as its vote. The PS collects all the votes, and counts the votes for all the positions to observe the most voted $K$ positions, which form the global sparsity mask. Hence, the sparsity mask can be written as
\begin{equation}
\mathbf{M}_{t}= S_{top}\left(\sum_{n=1}^N S_{top}\left(\vert\mathbf{g}_{n,t}\vert, K \right), K \right).
\end{equation}
Once the mask $\mathbf{M}_{t}$ is constructed, it is communicated to all the workers. Each worker sends a sparsified version of its local gradient estimate using the global mask, i.e., worker $n$ sends $\mathbf{g}_{n,t}\otimes\mathbf{M}_{t}$ to the PS to be aggregated. We call this scheme {\em sparse SGD with majority voting (SSGD-MV)}. 

We want to highlight that, although we consider a two-step process, the total number of transmitted bits in the uplink direction is the same with top-$K$ sparsification, and less number of bits are transmitted in the downlink direction. Later in Section \ref{section:SSGD_TCMV}, we will introduce a more efficient voting mechanism by utilizing the correlation among the gradient estimates over time, so that the number of transmitted bits in the uplink direction during the voting stage can be reduced.

\indent Before giving the detailed analysis of the  SSGD-MV scheme, we note that the proposed strategy can also be employed in the local SGD framework, in which each worker performs multiple local updates before communicating with the PS. Hence, we consider the general setup, where each worker performs $H$ local iterations, then sends the corresponding local model difference to the PS for aggregation and model update. The model difference of the $n$th worker at iteration $t$ can be written as 

\begin{equation} \label{model_update_mult}
\Delta\boldsymbol{\theta}_{n,t}=\sum^{H}_{h=1}-\eta_{t}\mathbf{g}^{(h)}_{n,t}, 
\end{equation}
where
\begin{equation}
\mathbf{g}^{(h)}_{n,t}=\nabla_{\boldsymbol{\theta}}F_{n}(\boldsymbol{\theta}^{(h)}_{n,t},\zeta_{n,i}),
\end{equation}
and
\begin{equation}
\boldsymbol{\theta}^{(h)}_{n,t}=\boldsymbol{\theta}^{(i-1)}_{n,t}-\eta_{t}\mathbf{g}^{(h)}_{n,t}.
\end{equation}
Here, we have set $\boldsymbol{\theta}^{(0)}_{n,t}=\boldsymbol{\theta}_{t}$.

The overall procedure of SSGD-MV is illustrated in Algorithm \ref{alg:SSGD_MV}. It has been shown that error accumulation/feedback technique combined with sparsification achieves better convergence results \cite{SGD.sparse2, SGD.feedback2, SGD.feedback3, SGD.feedback_double1, SGD.feedback_double2}. Hence, we employ the error accumulation strategy such that worker $n$ keeps track of the sparsification error, $\mathbf{e}_{n,t-1}$ at iteration $t$, to be aggregated to the local model difference $\boldsymbol{\Delta}_{n,t}$ of the next iteration as illustrated in line 9 and line 22 of Algorithm \ref{alg:SSGD_MV}.

\begin{algorithm}[ht]
\caption{Sparse SGD with Majority Voting}\label{alg:SSGD_MV}
\begin{algorithmic}[1]
    \FOR{$t=1,\ldots,T$}
        \STATE \underline{\textbf{Voting phase worker side:}}
        \FOR{$n=1,\ldots,N$}
            \STATE Receive $\widetilde{\boldsymbol{\Delta}}_{t-1}$ from PS
            \STATE \textbf{Update model:} $\boldsymbol{\theta}_{t} = \boldsymbol{\theta}_{t-1} + \widetilde{\boldsymbol{\Delta}}_{t-1}$
            \STATE initialize $\boldsymbol{\theta}^{(0)}_{n,t}=\boldsymbol{\theta}_{t}$
            \STATE \textbf{Perform $H$ local SGD steps:} 
            \STATE $\mathbf{\Delta}_{n,t}=\boldsymbol{\theta}^{(H)}_{n,t}-\boldsymbol{\theta}^{(0)}_{n,t}$
            \STATE{$\bar{\boldsymbol{\Delta}}_{n,t}=\mathbf{\Delta}_{n,t}+\mathbf{e}_{n,t}$}
            \STATE $\mathbf{M}_{n,t}=S_{top}(\vert\bar{\boldsymbol{\Delta}}_{n,t}\vert,K)$
            \STATE Send $\mathrm{supp}(\mathbf{M}_{n,t})$ to PS
        \ENDFOR
        \STATE\underline{\textbf{Voting phase PS side:}}
        \STATE{Apply majority voting:}
        \STATE {\color{blue}$\mathbf{M}_{t}=S_{top}(\sum^{N}_{n=1}\mathbf{M}_{n,t},K)$}
        \STATE {\color{red}$\mathbf{M}_{t}=S_{rand-weighted}(\sum^{N}_{n=1}\mathbf{M}_{n,t},K)$}
        \STATE{Send $\mathrm{supp}(\mathbf{M}_{t})$ to workers}
        \STATE \underline{\textbf{Sparse communication worker side:}}
        \FOR{$n=1,\ldots,N$}
        \STATE{$\widetilde{\mathbf{\Delta}}_{n,t}=\mathbf{M}_{t}\otimes\bar{\mathbf{\Delta}}_{n,t}$}
        \STATE Send $\widetilde{\mathbf{\Delta}}_{n,t}$ to PS
        \STATE $\mathbf{e}_{n,t+1}= \bar{\mathbf{\Delta}}_{n,t}-\widetilde{\mathbf{\Delta}}_{n,t}$
        \ENDFOR
    \STATE\underline{\textbf{Sparse communication PS side:}}
    \STATE{Aggregate local sparse gradients:}
    \STATE $\widetilde{\mathbf{\Delta}}_{t}=\sum_{n\in[N]}\widetilde{\mathbf{\Delta}}_{n,t}$
    \STATE send $\widetilde{\mathbf{\Delta}}_{t}$ to all workers
    \ENDFOR
\end{algorithmic}
\end{algorithm}

In Algorithm \ref{alg:SSGD_MV}, we consider two possible strategies to decide the winning locations of the voting procedure; the first one (highlighted with {\color{blue} blue}) simply counts the received votes from the workers and selects the top-$K$ most voted entries. Instead, in the second strategy, called majority voting with random selection (SSGD-MV-RS) (highlighted with {\color{red} red}), the winning locations are sampled randomly according to a distribution whose probabilities are proportional to the received votes. We note that the motivation behind the random selection strategy is to reduce any bias, similarly to the designs in \cite{SGD.sparse.rtopk, SGD.sparse.var}.

Next, we analyze the communication load of the proposed SSGD-MV strategy.

\subsection{Communication Load Analysis}
To measure the communication load, we use the total number of transmitted bits per iteration per users, denoted by $Q$. We note that $Q$ can be written as the sum of two quantities, $Q^{up}$ and $Q^{down}$, which correspond to uplink and downlink communication loads, respectively. Further, both $Q^{up}$ and $Q^{down}$ are equal to the sum of two terms, where the first one, $Q^{up}_{loc}/Q^{down}_{loc}$, corresponds to the number of bits used for majority voting to identify the non-zero positions, while the second one, $Q^{up}_{val}/Q^{down}_{val}$, is the number of bits used to transmit corresponding values.

Given the sparsity parameter $K=\phi \cdot d$, one can observe that
\begin{equation}Q^{up}_{val}=Q^{down}_{val}=q \phi d,
\end{equation}
where $q$ is the number of bits to represent each value according to the chosen quantization framework. Its default value is $q=32$, corresponding to 32 bits floating point precision. 

For the voting phase, each non-zero position can be represented with $log(d)$ bits, in the most naive approach, which implies $Q^{up}_{loc}=Q^{down}_{loc}=d\phi \log d$. However, as we later illustrate in Section \ref{ss:sparse_quant}, the total number of bits required to represent the positions of non-zero locations can be reduced to
\begin{equation}
Q^{up}_{loc}=Q^{down}_{loc}=d\phi(\log(1/\phi)+2).
\end{equation}
Consequently, the number of bits transmitted in both the uplink and downlink directions is given by
\begin{equation}\label{commload}
Q^{up}=Q^{down}= \phi d(\log(1/\phi)+ q + 2) .
\end{equation}

One can easily observe from (\ref{commload}) that, depending on the choice of the sparsification ratio $\phi$ and the number of bits to represent each model difference value $q$, transmission of the non-zero positions or the non-zero values may become the bottleneck. In other words, if the precision of the quantization, thus the number of required bits $q$, is reduced, then the voting mechanism will become the bottleneck in the communication load, thus the impact of the quantization technique will be limited by the voting phase. For instance, when $\phi=10^{-2}$, $9$ bits are required to represent the position of each non-zero value; therefore, when $q<9$, the communication load will be dominated by the voting phase. 

Hence, in order to utilize the quantization technique effectively, there is a need for a more efficient voting mechanism, particularly in the uplink direction. To this end, we introduce a more efficient voting mechanism in the next subsection, which  builds upon the assumption that the most important NN parameters, those demonstrating more visible changes in their values, do not change radically through the training process, which, in our setup, implies that the votes of the workers will be correlated temporally.

\section{Majority Voting with Time Correlation}\label{section:SSGD_TCMV}

In this section, we introduce a new voting mechanism to construct the sparsity mask, called majority voting with add-drop (SSGD-MV-AD), which builds upon the  aforementioned heuristic that  locations of the NN parameters that exhibit larger variations are correlated over iterations. Hence, our objective is to modify the voting process in order to utilize this correlation and reduce the required number of bits that must be communicated in the voting phase.

\begin{algorithm}[h]
\caption{Majority Voting with Add-Drop (MV-AD)}\label{alg:SSGD_TCMV}
\begin{algorithmic}[1]
    \FOR{$t=1,\ldots,T$}
        \STATE \underline{\textbf{Voting phase worker side:}}
        \FOR{$n=1,\ldots,N$}
        \STATE Receive $\widetilde{\mathbf{\Delta}}_{t-1}$ from PS
        \STATE  \textbf{Update model:} $\boldsymbol{\theta}_{t} = \boldsymbol{\theta}_{t-1} + \widetilde{\boldsymbol{\Delta}}_{t-1}$
        \STATE  initialize $\boldsymbol{\theta}^{(0)}_{n,t}=\boldsymbol{\theta}_{t}$
        \STATE  \textbf{Perform local SGD for $H$ iteartions:} 
        \STATE  $\boldsymbol{\Delta}_{n,t}=\theta^{(H)}_{n,t}-\theta^{(0)}_{n,t}$
        \STATE {$\bar{\boldsymbol{\Delta}}_{n,t}=\boldsymbol{\Delta}_{n,t}+\mathbf{e}_{n,t}$}
        \STATE  $\mathbf{M}_{n,t}=S_{top}(\vert\bar{\boldsymbol{\Delta}}_{n,t}\vert,K)$
        \STATE  \textbf{Add-drop mechanism:} 
    \STATE   $\mathbf{M}^{add}_{n,t}$=
    \STATE  $S_{top}(\max((\mathbf{M}_{n,t}-\widetilde{\mathbf{M}}_{n,t-1}),0)\otimes\bar{\boldsymbol{\Delta}}_{n,t},K_{ad})$ 
    \STATE   $\mathbf{M}^{drop}_{n,t}=$
     \STATE  $S_{bot}(\max((\widetilde{\mathbf{M}}_{n,t-1}-\mathbf{M}_{n,t}),0)\otimes\bar{\boldsymbol{\Delta}}_{n,t},K_{ad})$ 
     \STATE  $\widetilde{\mathbf{M}}_{n,t}=\widetilde{\mathbf{M}}_{n,t-1}+\mathbf{M}^{add}_{n,t}-\mathbf{M}^{drop}_{n,t}$
    \ENDFOR
\STATE \underline{\textbf{Voting phase PS side:}}
\STATE {Apply voting}
\STATE  $\mathbf{M}^{sum}_{t}=\mathbf{M}^{sum}_{t-1} + \sum^{N}_{n=1}( \mathbf{M}^{add}_{n,t}-\mathbf{M}^{drop}_{n,t})$
\STATE {Count the votes:}
\STATE  $\mathbf{M}_{t}=S_{top}(\mathbf{M}^{sum}_{t},K)$
\STATE {Send $\mathbf{M}_{t}$ to workers}
\STATE  \underline{\textbf{Sparse communication worker side:}}
     \FOR{$n=1,\ldots,N$}
      \STATE {$\widetilde{\mathbf{\Delta}}_{n,t}=\mathbf{M}_{t}\otimes\bar{\boldsymbol{\Delta}}_{n,t}$}
        \STATE  Send $\widetilde{\mathbf{\Delta}}_{n,t}$ to PS
        \STATE  $\mathbf{e}_{n,t}= \bar{\boldsymbol{\Delta}}_{n,t}-\widetilde{\mathbf{\Delta}}_{n,t}$
\ENDFOR
\STATE \underline{\textbf{Sparse communication PS side:}}
\STATE {Aggregate local sparse model updates:}
\STATE  $\widetilde{\boldsymbol{\Delta}}_{t}=\sum_{n\in[N]}\widetilde{\boldsymbol{\Delta}}_{n,t}$
    \STATE  send $\widetilde{\boldsymbol{\Delta}}_{t}$ to all workers
    \ENDFOR
\end{algorithmic}
\end{algorithm}

To this end, we propose an \textit{add-drop} mechanism for the voting phase. In the original majority voting mechanism, each worker sends the locations of the largest $K$ absolute values, where $K$ corresponds to the $\phi$ portion of the total gradient dimension. When the add-drop mechanism is employed, the current vote of each worker is obtained by changing only a certain portion of the its previous vote. Hence, at each iteration, each worker sends the list of indices to be removed/added to the previously voted indices. We use $K_{ad}$ to denote the  maximum number of changes allowed during voting; that is, at each iteration at most $K_{ad}$ number of indices can be removed/added to the voting list by each worker, which corresponds to $\phi_{ad} << \phi$ portion of all the parameters.

\indent Let $\widetilde{\mathbf{M}}_{n,t}$ denote the votes of worker $n$ at iteration $t$ according to the add-drop mechanism, which can be executed in the following way: to obtain the indices to be added at iteration $t$, worker $n$ identifies the largest $K_{ad}$ absolute values in the current top-$K$ values $\mathbf{M}_{n,t}$ that are not present in the previous vote $\widetilde{\mathbf{M}}_{n,t-1}$ (as illustrated in line 14 of Algorithm \ref{alg:SSGD_TCMV}), and similarly, to obtain the indices to be removed at iteration $t$, worker $n$ identifies the $K_{ad}$ lowest absolute values in the previous vote $\widetilde{\mathbf{M}}_{n,t-1}$ that are not in the current top-$K$ values $\mathbf{M}_{n,t}$ (as illustrated in line 12 of the algorithm \ref{alg:SSGD_TCMV}).

\indent We remark that the PS does not need to keep track of the previous votes of the all workers. As illustrated in Algorithm \ref{alg:SSGD_TCMV} (line 20), it is sufficient to keep track of the cumulative sum of the votes $\mathbf{M}^{sum}_{t}$, which is updated according to the following rule:
\begin{equation}
\mathbf{M}^{sum}_{t}=\mathbf{M}^{sum}_{t-1} + \sum^{N}_{n=1}( \mathbf{M}^{add}_{n,t}-\mathbf{M}^{drop}_{n,t}).
\end{equation}
\indent If we revisit our example scenario where $\phi=10^{-2}$ and set $\phi_{ad}=10^{-3}$, which corresponds to assuming $90\%$ correlation over time, the number of bits required to represent each position will now be 12 bits, but the number of transmitted positions will be reduced by a factor of $5$; hence, effectively the number of bits per position will be only 2.4.

\indent We want to highlight that, although the main design goal of the add-drop mechanism is to reduce the communication load of the voting mechanism, in parallel it also functions similarly to the network pruning strategy. In a sense, it naturally seeks a sparse version of the NN architecture since the add-drop mechanism enforces training along a direction dictated by a small subset of the parameters.

\section{Numerical Results}\label{section:numerical}
\subsection{Simulation Setup}
\indent To evaluate the performance of the proposed majority voting strategies, we consider the image classification task on the CIFAR-10 dataset \cite{cifar-10}, which consists of 10 RGB image classes, and 50K training and 10K test images, respectively. For the training, we choose ResNet-18 neural network architecture \cite{NN.DRN}, which consists of 8 basic blocks, each with two 3x3 convolutional layers and batch normalization. After two consecutive basic blocks, image size is halved with an additional 3x3 convolutional layer employing stride. This network consists of 11,173,962 trainable parameters altogether. We consider $N=10$ workers, and a federated setup, where the available training dataset is divided among the workers in a disjoint manner. The images, based on their classes, are distributed in an identically and independently distributed (IID) manner among the workers.\\

\subsection{Simulation Scenario}\label{ss:scenario}
\begin{figure*}[h]
\begin{center}
\includegraphics[scale =0.4]{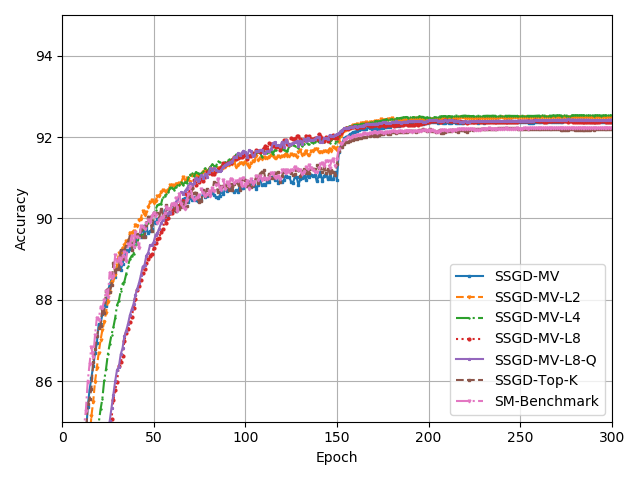}
\caption{Test accuracy results over 300 epochs for SSGD-MV variations, SSGD-top-$K$ and SM benchmark}
\label{300_epoch_MV_only}
\end{center}
\end{figure*}

\begin{figure*}[h]
\begin{center}
\includegraphics[scale =0.4]{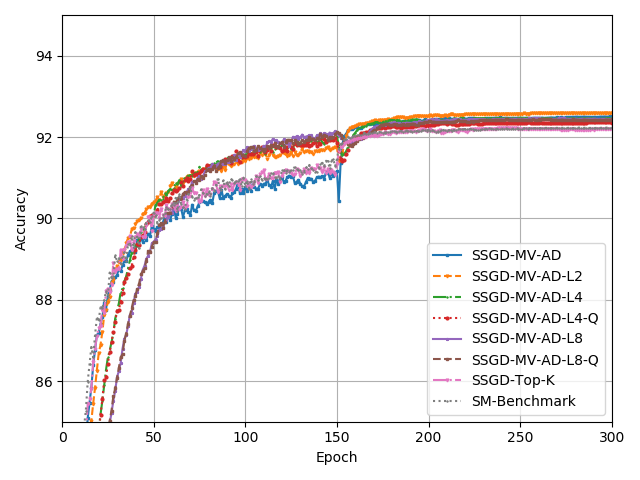}
\caption{Test accuracy results over 300 epochs for SSGD-MV-AD variations, SSGD-top-$K$ and SM benchmarks.}
\label{300_epoch_mv_ad}
\end{center}
\end{figure*}

\begin{figure*}[h]
\begin{center}
\includegraphics[scale =0.38]{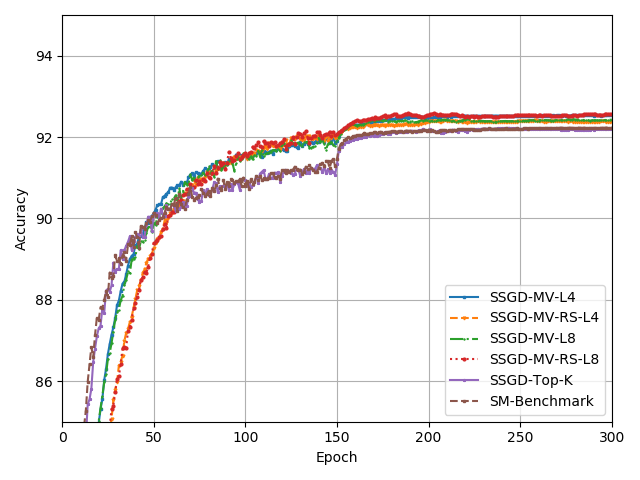}
\caption{Test accuracy results over 300 epochs for SSGD-MV,SSGD-MV-RS, SSGD-top-$K$ and the SM benchmark.}
\label{300_epoch_mv_random}
\end{center}
\end{figure*}

\begin{figure*}[t!]
    \centering
    \begin{subfigure}[t]{0.47\textwidth}
        \centering
        \includegraphics[height=2.4in]{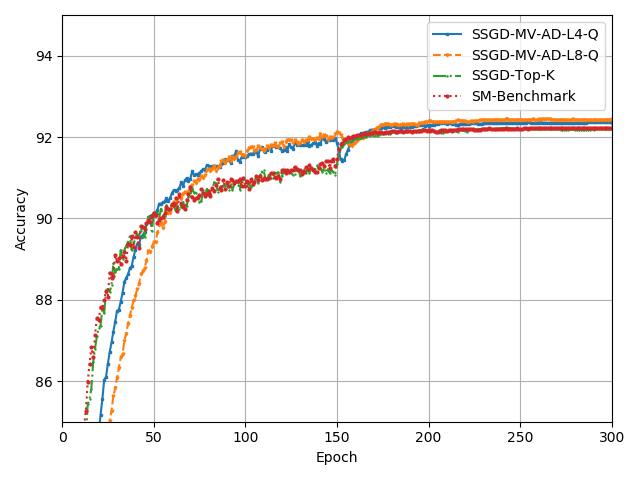}
        \caption{Comparison of the SSGD-MV variations, in high compression regime, and the benchmark schemes.}
        \label{QuantizedFig}
    \end{subfigure}%
    ~ 
    \begin{subfigure}[t]{0.47\textwidth}
        \centering
        \includegraphics[height=2.4in]{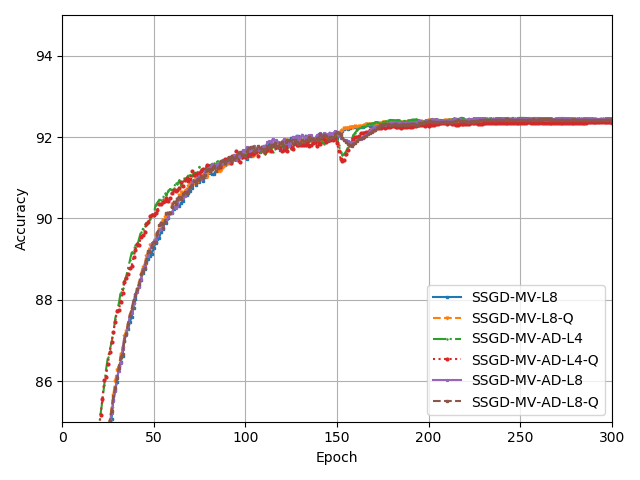}
        \caption{Comparison of the test accuracy results of SSGD-MV strategies with and without quantization.}
        \label{QuantizedvsUnQfig}
    \end{subfigure}
    \caption{Test accuracy results over 300 epochs when quantization is employed.}
    \label{quantizedExperiments}
\end{figure*}

\begin{table*}[ht]
 \caption{Top-1 test accuracy and the compression rate of the studied schemes. Schemes with the highest test accuracy and the highest compression rate are highlighted in \textbf{bold}.}
 \label{top1_acc_table}
\begin{tabular}{ | M{3.5cm} | M{2cm}| M{3cm}| M{3cm}| M{3cm}|}
\toprule
    \textbf{Method} & \textbf{Top-1 Accuracy (mean $\pm$ std)} & \textbf{Uplink bit budget} & \textbf{Downlink bit budget} &  \textbf{Compression Rate Uplink/ Downlink}\\ 
\midrule
    SSGD-MV & 92.36 $\pm$ 0.21 & $\bar{q}_{loc} = 9\times10^{-2}$  $\bar{q}_{val} = 3.2\times10^{-1}$ & $\bar{q}_{loc} = 9\times10^{-2}$,  $\bar{q}_{val} = 3.2\times10^{-1}$ &$\times78$ / $\times78$\\
    SSGD-MV-L2 & 92.5 $\pm$ 0.15 & $\bar{q}_{loc} = 4.5\times10^{-2}$,  $\bar{q}_{val} = 1.6 \times10^{-1}$ & $\bar{q}_{loc} = 4.5\times10^{-2}$,  $\bar{q}_{val} = 1.6 \times10^{-1}$ & $\times156$ / $\times156$\\
    SSGD-MV-L4 & 92.54 $\pm$ 0.27 & $\bar{q}_{loc} = 2.25\times10^{-2}$,  $\bar{q}_{val} = 8 \times10^{-2}$ & $\bar{q}_{loc} = 2.25\times10^{-2}$,  $\bar{q}_{val} = 8 \times10^{-2}$ & $\times312$ / $\times312$\\
    SSGD-MV-L8 & 92.37 $\pm$ 0.17 & $\bar{q}_{loc} = 1.125\times10^{-2}$,  $\bar{q}_{val} = 4 \times10^{-2}$ & $\bar{q}_{loc} = 1.125\times10^{-2}$,  $\bar{q}_{val} = 4 \times10^{-2}$ & $\times624$ / $\times624$\\
    SSGD-MV-L8-Q & 92.43 $\pm$ 0.14 & $\bar{q}_{loc} = 1.125\times10^{-2}$,  $\bar{q}_{val} = 5 \times10^{-3}$ & $\bar{q}_{loc} = 1.125\times10^{-2}$,  $\bar{q}_{val} = 4 \times10^{-2}$ & $\times2000$ / $\times624$\\
    SSGD-MV-RS-L4 & 92.43 $\pm$ 0.2 & $\bar{q}_{loc} = 2.25\times10^{-2}$,  $\bar{q}_{val} = 8 \times10^{-2}$ & $\bar{q}_{loc} = 2.25\times10^{-2}$,  $\bar{q}_{val} = 8 \times10^{-2}$ & $\times312$ / $\times312$\\
    SSGD-MV-RS-L8 & 92.57 $\pm$ 0.24 & $\bar{q}_{loc} = 2.25\times10^{-2}$,  $\bar{q}_{val} = 8 \times10^{-2}$ & $\bar{q}_{loc} = 2.25\times10^{-2}$,  $\bar{q}_{val} = 8 \times10^{-2}$ & $\times624$ / $\times624$\\
    SSGD-MV-AD & 92.5 $\pm$ 0.19 & $\bar{q}_{loc} = 2.4\times10^{-2}$  $\bar{q}_{val} = 3.2\times10^{-1}$ & $\bar{q}_{loc} = 9\times10^{-2}$,  $\bar{q}_{val} = 3.2\times10^{-1}$ &$\times93$ / $\times78$\\
    \textbf{SSGD-MV-AD-L2} & \textbf{92.6} $\pm$ \textbf{0.19} & $\bar{q}_{loc} = 1.2\times10^{-2}$  $\bar{q}_{val} = 1.6\times10^{-1}$ & $\bar{q}_{loc} = 4.5\times10^{-2}$,  $\bar{q}_{val} = 1.6\times10^{-1}$ &$\times186$ / $\times156$\\
    SSGD-MV-AD-L4 & 92.46 $\pm$ 0.22 & $\bar{q}_{loc} = 6\times10^{-3}$  $\bar{q}_{val} = 8\times10^{-2}$ & $\bar{q}_{loc} = 2.25\times10^{-2}$,  $\bar{q}_{val} = 8\times10^{-2}$ &$\times372$ / $\times312$\\
    SSGD-MV-AD-L4-Q & 92.36 $\pm$ 0.19 & $\bar{q}_{loc} = 6\times10^{-3}$  $\bar{q}_{val} = 1\times10^{-2}$ & $\bar{q}_{loc} = 2.25\times10^{-2}$,  $\bar{q}_{val} = 8\times10^{-2}$ &$\times2000$ / $\times312$\\
    SSGD-MV-AD-L8 & 92.45 $\pm$ 0.2 & $\bar{q}_{loc} = 3\times10^{-3}$  $\bar{q}_{val} = 4\times10^{-2}$ & $\bar{q}_{loc} = 1.125\times10^{-2}$,  $\bar{q}_{val} = 4\times10^{-2}$ &$\times745$ / $\times624$\\
    \textbf{SSGD-MV-AD-L8-Q} & 92.43 $\pm$ 0.23 & $\bar{q}_{loc} = 3\times10^{-3}$  $\bar{q}_{val} = 5\times10^{-3}$ & $\bar{q}_{loc} = 1.125\times10^{-2}$,  $\bar{q}_{val} = 4\times10^{-2}$ &$\boldsymbol{\times4000}$ / $\boldsymbol{\times624}$\\
    SSGD-top-$K$  & 92.194 $\pm$ 0.324 & $\bar{q}_{loc} = 9\times10^{-2}$  $\bar{q}_{val} = 3.2\times10^{-1}$ & $\bar{q}_{loc} = 6\times10^{-1}$  $\bar{q}_{val} = 3.2$ & $\times78$ / $\times8.4$\\
    SM-Benchmark & 92.228 $\pm$ 0.232 & - & -  & -\\
  \bottomrule  
\end{tabular}
\end{table*}

We have implemented the proposed three novel algorithms, and tested them with different numbers of local gradient iterations and quantization methods. These algorithms are i) SSGD-MV, ii) SSGD-MV-AD, and iii) SSGD-MV-RS. To compare with these algorithms we also implemented sparse SGD with top-$K$ batch and single machine (SM) benchmark. Each variation of the simulation results is an average of 10 separate trails, apart from SSGD-MV-RS, whose results are averaged across 5 separate trials for each of its own variations.

In the presentation of our numerical results, we denote the number of local steps $H$ with L$H$, and if we apply quantization to the algorithm, we denote it only by adding a Q symbol, as we only utilize 4 bit quantization. For example, SSGD-MV-L4-Q stands for quantized sparse SGD with majority voting with $H=4$ local iterations.\\

In SSGD-MV, we took the indices of the top $\phi=10^{-2}$ absolute model difference values from the PS model, after the $H$-th local model update, where we experimented with $H=2,4,8$. Each worker creates a mask based on its top $\phi=10^{-2}$ parameters, and sends its support to the parameter server which is where we apply majority voting system to make a global mask which is used by all workers to send the masked selected parameters.

Unlike SSGD-MV, where each worker votes for the ratio of $\phi = 10^{-2}$ parameters, in SSGD-MV-AD, workers only make $10\%$ change in their previous vote; specifically $\phi_{add}=10^{-3}$ to add new voted parameters and $\phi_{drop}=10^{-3}$ remove from the previous vote. Workers make those changes at the end of $H$-th local iteration. When all workers communicate their masks, parameter server computes the new global mask by utilizing majority voting system.\\

In SSGD-MV-RS, the parameter server calculates the global mask by randomly selecting the indices using the votes until $\phi=10^{10-2}$ number of unique indices are selected. Due to the algorithm's nature, if a certain index is voted by an overwhelming majority, it has a higher chance of being selected for the global mask. \\

We set starting learning rate to $\eta = 0.5$ for all the experiments aside of the SM benchmark which we set starting learning rate to $\eta = 0.1$.

\indent In all the implementations we employ the warm up strategy \cite{large_scale_training}, where the learning rate is initially set to $\eta=0.1$, and is increased to its corresponding scaled value gradually in the first 5 epochs. We also note that, we do not employ the sparsification and quantization methods during the warm up phase.\\

\indent In all the simulations, the DNN architecture is trained for 300 epochs and the learning rate is reduced by a factor of 10 after the first 150 and 225 epochs, respectively \cite{NN.DRL, NN.DRN}. Lastly, in all the simulations we employ L2 regularization with a given weight decay parameter $10^{-4}$.\\

\subsection{Sparse and Quantized Representation}\label{ss:sparse_quant}
Here, we will briefly explain how we represent the non-zero positions of a sparse vector and its quantized values.

\subsubsection{Sparse representation} 

The position of each non-zero value can be represented using $\log(d)$ bits. However, as the size of the DNN architectures are taken into account, with millions of parameters, this means more than 20 bits for each non-zero position. Instead, we propose an alternative method to represent the non-zero positions. in which the number of bits per each nonzero location is not a parameter of $d$, but the sparsification ratio $\phi$. 

For a given sparse vector $\mathbf{g}$ with a sparsity ratio of $\phi$, the proposed sparse representation strategy works in the following way. First, $\mathbf{g}$ is divided into blocks of size $1/\phi$, which creates $d\phi$ equal-length blocks. Within each block, the positions of non-zero values can be represented using $\log(1/\phi)$ bits. Now, to complete the sparse representation of the vector, we need an identifier for the end of each block, hence we use a one bit symbol, i.e., $0$, to indicate the end of block. Accordingly, we append a one bit symbol, i.e., $1$, to the beginning of each $\log(1/\phi)$ bits used for the intra-block positioning.

\indent Once the sparse representation is sent, receiver starts reading from the beginning of the bit stream and if the initial bit is $1$, then reads the next $\log(1/\phi)$ bits to recover the intra-block position and move the cursor accordingly, if the initial bit is $0$, then updates the block index and moves the cursor accordingly, this process is continued until the cursor reaches the end of the bit stream. From the communication aspect, on the average, $\log (1/\phi) + 2$ bits are used for position of the each non-zero value. To clarify with an example, when $\phi\in[1/64,1/128]$, 9 bits are sufficient for the sparse representation, which is less than half of the required bits for the naive sparse representation, when DNN architectures are considered.


\subsubsection{Quantized Representation}

For the quantized representation we consider a framework building  upon the scaled sign operator \cite{SGD.feedback2,SGD.q0}, which returns the quantized representation of a vector $\mathbf{v}$ as
\begin{equation}
\mathcal{Q}(\mathbf{v})=\vert\vert \mathbf{v}\vert\vert_{1}/d\cdot \mathrm{sign}(\mathbf{v}),
\end{equation}
where $d$ is the dimension of vector $\mathbf{v}$. Further, it has been shown that when the vector is divided into smaller blocks and  quantization is applied to each block separately, it is possible to reduce the quantization error further \cite{SGD.q0}. Inspired by the natural compression approach in \cite{SGD.q9}, and the aforementioned blockwise quantization strategy in \cite{SGD.q0}, we employ a fractional quantization strategy. The fractional quantization scheme works as follows:  first, it identifies the maximum and minimum values in $\vert\mathbf{v}\vert$, denoted by $v_{max}$ and $v_{min}$, respectively, then for the given bit budget $q$ the interval $[v_{max},v_{min}]$ is divided into $L=2^{q-1}$ subintervals $I_{1},\ldots,I_{L}$ such that
\begin{equation}
I_{l}=\left[\frac{v_{\max}}{\alpha^{l-1}}, \frac{v_{\max}}{\alpha^{l}}\right],
\end{equation}
where $\alpha=(v_{\max}/v_{\min})^{1/L}$.

Hence, for the quantized representation of each value in $\vert\mathbf{v}\vert$, we use index of the corresponding interval using $q-1$ bits and one more bit for the sign, thus $q$ bits in total. After forming a bitstream of the quantized representation we append $L$ mean values $\mu_{1}, \ldots, \mu_{L}$, to the end, where $\mu_{l}$ is the average of the values in $\vert\mathbf{v}\vert$, which falls into the subinterval $I_{l}$. Hence, the recovered values at the receiver side takes values from the set  $\left\{\mu_{1},\ldots,\mu_{L}\right\}$.

\subsection{Simulation Results}
As explained in Section \ref{ss:scenario}, in total we consider three different variations of the introduced majority voting strategy with 13 different implementations in total. We provide all the corresponding top-1 test accuracy results and the achieved compression rates both in the uplink and downlink direction in Table ~\ref{top1_acc_table}. To provide a better understanding of the compression rates given in Table ~\ref{top1_acc_table}, we also provide the effective bit budgets used for representing the quantized values and the non-zero positions of the sparse vector, which are denoted by $\bar{q}_{val}$ and $\bar{q}_{val}$ respectively. We note that $\bar{q}_{val}$ and $\bar{q}_{loc}$ are equivalent to  $Q_{val}/dH$ and $Q_{loc}/dH$, respectively.

In Figures \ref{300_epoch_MV_only} and \ref{300_epoch_mv_ad}, we illustrated the effects of the local steps for SSGD-MV and SSG-MV-AD, respectively. From these figures we can see that the benefits of local SGD steps are not limited to the communication load; but it also helps with the convergence as the local step number reaching higher $H$ values. This is apparent for SSGD-MV in Figure \ref{300_epoch_MV_only}. but it is most prominent for the SSGD-MV-AD by preventing the accuracy drops at epoch 150 which where we decay the starting learning rate for the first time as shown at Figure \ref{300_epoch_mv_ad}.
Aside from the healthier convergence, Table ~\ref{top1_acc_table} shows that local steps of $H=2,4$ also benefits to the consistency of the results by achieving lower standard deviation in their own algorithm types and also compression rate is multiply by $2$ with each consecutive even local steps.
In Figure \ref{300_epoch_mv_random}, we observe that  SSGD-MV-RS performs similarly to the SSGD-MV, although still outperforms the benchmark algorithms.

In Figures \ref{quantizedExperiments}, we monitor the impact of quantization on the test accuracy of the SSGD-MV strategy. In Figure \ref{QuantizedvsUnQfig}, we plot the test accuracy results of the 3 SSGD-MV variations namely SSGD-MV-L8, SSGD-MV-AD-L4, SSGD-MV-AD-L8, and their quantized versions. We observe from the figure that the quantization has a very minor impact on the test accuracy, although it reduces the number of bits used for representing the each non-zero value by factor of 8. In Figure \ref{QuantizedFig}, we compare the two majority voting based schemes that achieve the highest compression rates namely, SSGD-MV-AD-L4-Q and  SSGD-MV-AD-L8-Q, with the two benchmark schemes to demonstrate that even with high compression rates majority voting strategy outperforms the benchmark schemes.

From the simulation results, we observe that SSGD-MV-AD-L2 achieves the highest test accuracy, which may seem counter-intuitive at first glance, from the communication load- accuracy trade-off perspective, since both add-drop mechanism and local steps are used to reduce the communication load. However, it has been shown that local SGD may achieve higher performance compared to the mini-batch alternative \cite{SGD.local2, SGD.localx}. Besides, the use of local steps increases the accuracy of the voting mechanism since the variations in the values over $H$ local steps provides more reliable information for identifying the most important positions compared to the single gradient estimate, due to the randomness in the stochastic gradient computations. This interpretation is also supported by the comparison of the test accuracy results of SSGD-MV, SSGD-MV-L2, SSGD-MV-L4, and SSGD-MV-L8, where the highest test accuracy is achieved by SSGD-MV-L4. These observations are also consistent with the previous findings on the performance of local SGD, where the test accuracy first increases with the number of local steps $H$, but then starts to decline after a certain point. We also note that one disadvantage of using larger $H$ is causing delay in the error accumulation, so that the error aggregated for the local difference becomes outdated.\\

\indent According to the simulation results, SSGD-MV-AD-L2 outperforms  SSGD-MV-L2. We argue that, as briefly discussed in Section \ref{section:SSGD_TCMV}, the introduced add-drop mechanism, in addition to reducing the communication load, also acts as a natural regularizer for NN training, and thus serves to achieve better generalization. As we already mentioned above, the use of local steps improves the performance of the voting mechanism, which explains the difference between the accuracy of SSGD-MV-AD-L2 and SSGD-MV-AD.\\

\indent In our simulations, we also combine the proposed majority voting based sparsification approach with the quantization strategy to reduce the communication load further and observe that it is possible to achieve up to $\times4000$ compression in total, using SSGD-MV-AD-L8-Q4, without a noticeable loss in the test accuracy compared to the SM benchmark.\\
\indent Finally, we want to remark that the achieved accuracy results can be improved further by employing accelerated SGD variants, such as momentum SGD or/and applying layer-wise sparsification instead of using a global one \cite{SGD.sparse0}. We will consider these extensions as future work.

\section{Conclusion}
In this paper, we introduced a majority voting based sparse communication strategy for the DSGD framework. The proposed majority voting strategy, particularly the one enriched with the add-drop mechanism, has three main benefits: First, it reduces the communication load further by taking advantage of the correlation among the model updates across iterations. Second, it serves as a natural regularization on the NN sparsity to achieve better generalization, and finally it works as a linear operator. By conducting extensive simulations, we showed that the proposed SSGD-MV  strategy (and its variants) achieves the benchmark test accuracy of a  single machine. When combined with localization and quantization strategies, it can achieve impressive compression rates as high as $\times4000$.
\bibliography{ref}
\bibliographystyle{mlsys2020}

\end{document}